\documentclass[12pt]{article}
\usepackage{amsmath,amssymb,amsthm,amscd,pstricks}
\newcommand{\id}{\operatorname{Identity}}
\newcommand{\cst}{\operatorname{Cst}}
\newcommand{\uns}{\operatorname{Uns}}
\newcommand{\lin}{\operatorname{LFs}}
\newcommand{\abs}{\operatorname{Abs}} \newcommand{\tim}{t}
\newcommand{\pined}{\smallsetminus\{0\}}
\newcommand{\absolute}[1]{\lvert#1\rvert}
 \newcommand{\ul}{\underline}
\newcommand{\mma}{\emph{Mathematica}}
\newcommand{\downmapsto}{\rotateright{\mapsto}}
\newcommand{\downeq}{\rotateright{=}}
\newcommand{\new}[1]{\widetilde{#1}} \numberwithin{equation}{section}
\newtheorem{definition}{Definition}[section]
\newtheorem{theorem}{Theorem}[section]

\title{A theory of experiment}

\author{Pierre Albar\`ede\thanks{r\'es.  Valvert, 12 rue de la
    Fourane, 13090 Aix-en-P., France, pierre.albarede@free.fr}}

\date{\today}
\begin{document}
\maketitle

\begin{abstract}
  This article aims at clarifying the language and practice of
  scientific experiment, mainly by hooking observability on
  calculability.
   %
\end{abstract}
%

\section{Motivation}

Scientific knowledge is traditionally based on logic and experiment.
However, I will defend that experiment involves a lot of logic, if not
only logic.

As an experimentalist, I have found out, which is not easy to admit,
that experiments do fail logically as well as technically.
``Experiment, observable, error, interpretation, constant\ldots'' are
ill-defined.  Every experimental constant hides a potential variable.
Technology and jargon are often hopeless attempts to escape a logical
quagmire.  The very difficulty of discussing experiment indicates a
need of logic, where it has been too quickly eliminated, to explain
and prevent failures, and eventually to sharpen our understanding of
nature, including ourselves.

``Analog computing, numerical experiment, thought experiment,
experimental program, object programming, computing hardware, natural
or real number'' implicitly relate computing and experiment.  Could
the art of experiment develop somehow like the art of computer
programming, at least for the sake of their cooperation?  I will
defend that experiment is computing, by hooking observability on
calculability, a well-established mathematical concept
\cite{Delahaye}.

A theory of experiment should describe: -~objects, with just as much
detail as needed, symbolically, -~relations, between a user and
objects, -~interactions, between objects, -~the user's action, to
extract information from objects.  A duplicity of relation and action
is a feature of both experiment and functional programming.  Both
symbolism and functional programming are supported by \mma\ 
\cite{Wolfram, BarrereMma}.
  
\section{Formal system}

\subsection{Calculability}

The user expresses $y$ in a natural lexicon by $x$ in a formal
lexicon, $y\triangleright x,x\triangleleft y$.  The formal expresses
the natural: $\text{time}\triangleright\tim$, let $\tim$ be time, for
time $\tim$.  An object is nothing but what the user expresses, an
object of expression. An \emph{atom} is a symbol, like $\pi,z,\delta$,
or a numeral, like $3,22/7,3.14$; although $\pi$ expresses a number
(the half-circle perimeter), it is not a numeral.  The formal lexicon
is not outside the natural lexicon; on the contrary, as we speak
naturally of $\pi$, the natural tends to encompass the formal.

An expression immediately does or does not match a pattern (not a
set), expressed by $\_$. For example, $\_$ matches any expression,
$x\_$ matches any expression, to be named $x$,
$\_\operatorname{Integer}$ matches any integer numeral.

A formal system rewrites or evaluates any expression given by the
user, according to rules, possibly using patterns. Although an
expression may mean something for the user, meaning must not affect
evaluation. A rule is potential or delayed evaluation. For $y$ in the
formal lexicon, as there is no better expression of $y$ than $y$
itself, $y\triangleright x$ is expressed by a replacement rule,
$x\rightarrow y$, so that the formal system will evaluate $x\mapsto
y$. $\triangleright,\triangleleft,\mapsto$ are not in the formal
lexicon, we use them to speak about the user and the formal system;
they are meta-symbols (like meta-characters in a typesetting system).

A program is an assembly of rules (obeying meta-rules).  The user can
rule and interrupt evaluation.  Without him, the formal system does
nothing, or runs in circles: \textsc{homo ex machina} prevents
\textsc{in girum imus nocte}.  This also sounds like the principle of
inertia in mechanics.

I take for a definition of recursivity and calculability
\cite[3.1.2]{BarrereMma}.  Calculability means evaluating into
$\_\operatorname{Integer}$ or any other numerical pattern, like
$\_\operatorname{Real}$. For example, $\pi$ is calculable, but
$\Omega$ defined in \cite{Chaitin} is not.

An expression is -- like a tree, with a trunk, branches and leaves, or
a body, with a head, members and fingertips -- a headed sequence of
atoms or, recursively, expressions, like $x[y,z[1,2]]$ or $x+y$,
standing for $\operatorname{Plus}[x,y]$.  The depth of an expression
is its recursion number, plus one,
$\operatorname{Depth}[x[y,z[1,2]]]\mapsto 3$.

\begin{definition}
  A formal system is universal if it can be ruled to evaluate for any
  other formal system.
    \label{defUniForSys}
\end{definition}
A Turing machine is a universal formal system.  Many microprocessors
are Turing machines.  The complexity of an expression, whatever the
controversial definition of complexity \cite{Delahaye}, depends on the
formal system.
\begin{definition}
  For a formal system $f$, let $c[x,f]\in\mathbb{R}\cup\{\infty\}$ be
  the complexity of $x$ in $f$.  $c[x,f]=\infty$ if $x$ is
  non-calculable.  For two formal systems $f,g$, $f$ is specialized in
  $x$, more than $g$, if $c[x,f]\le c[x,g]$.
    \label{defSpecialization}
\end{definition}

Specialization is reusable complexity, packaged as a subroutine or a
black box, with public (non-private) rules, that are all what the user
needs to know.
\begin{itemize}
\item A microprocessor is specialized in floating-point arithmetic by
  a so-called floating-point unit.
    
\item Computational fluid dynamics aims at specializing a formal
  system in solving fluid-dynamics equations.  The public rules are
  boundary conditions and thermodynamic laws, like an equation of
  state.
    
\item A set of wires, resistors, generators and gauges can contribute
  to a formal system, specialized in solving linear algebraic
  equations with real coefficients.  The public rules are (Ohm and
  Kirchhoff) laws of electrokinetics.
    
\item A primitive idea of geometry is that a material system may be
  ruled to partly evaluate itself, as the universe is ruled by the
  gnomon to evaluate star positions \cite{SerresGnomon}.
\end{itemize}
Analog computing is when private rules are discovered, digital
computing is when private rules are invented.  However, the difference
between discovery and invention (or analysis and synthesis, after
Kant) is thin, because both are interwoven.  Anyway, the user is
mostly concerned by computing power, not the status of private rules.

Noise seems to be a particular drawback of analog computing.  However,
an analog system evaluating $x$ with noise can be considered as
evaluating the probability density function of $x$ without noise.  In
digital computing, apart from purposely introduced Monte-Carlo noise
and large thermodynamic fluctuations, floating point arithmetic makes
only round-up errors.  But round-up errors depend on random hardware,
precision, units and programming: for example, numerically ($N$),
$N[\log[3/2]]\neq N[\log[3]]-N[\log[2]]$, so that $\log$ can be
considered as noisy.  Finally, many techniques of numerical evaluation
are available, and no one is perfect.

\subsection{Logic and algebra}

The formal system should support not only replacement rules, but also
relational rules $\_!$ (not $\operatorname{Factorial}$), and evaluate
accordingly, as in
\[
x\in A!\ x\in A\mapsto\operatorname{True}.
\]

G\H{o}del's completude and incompletude theorems set limits on logical
evaluation, independently of technique: a formal system supporting the
rules of logic but not arithmetic ($\mathbb{N}$), or finite logic, can
evaluate all its relational expressions, while a formal system
supporting the rules of both logic and arithmetic, or infinite logic,
cannot.  Nevertheless, the user can always rule what a formal system
cannot evaluate, at his own risk of inconsistency or redundancy.

\mma\ partly supports some semantic patterns, like
$x\_/;x\in\mathbb{R}$, trying to match any expression of a real
number, whatever its syntax. \emph{Semantica} \cite{HarrisJ}, a \mma\ 
add-on, based on $\operatorname{Solve}$, supports more semantic
patterns, like $(2x\_)_{s}$, matching any expression with the double
of its half.  Semantic pattern matching makes \mma\ look like it
evaluates according to its own understanding of sets.  However,
internally, a semantic pattern reduces to syntactic patterns, covering
limited instances of the intended semantic pattern.  The incompletude
theorem implies that no formal system can completely support semantic
pattern matching \cite{HarrisJ}.  The ultimate responsibility for
semantic pattern matching belongs to the user.

Although an object is identical only to itself
\cite[5.5303]{WittgensteinTractatus}, the user may want to rule
identities between expressions of one object, as in
\begin{eqnarray}
    \pi&\triangleright&3,\ 3.14,\ \frac{22}{7},\
    (\sum_{n=1}^{\infty}\frac{6}{n^{2}})^{1/2}=
    2\operatorname{ArcSin}[1]=\oint\frac{d\theta}{2}=\pi. 
    \nonumber
\end{eqnarray}
$x=y$ means $x\triangleright y,y\triangleright x$.  The formal system
should be able to evaluate some identities, if not all.
$x=y$ legitimates both $x\rightarrow y,y\rightarrow x$, but, in
general, which one is more useful cannot be decided once for ever.  A
solution, suggested by Burindan's donkey, the wise animal, would be to
keep $x,y$ under the constraint $x=y$.

\begin{definition}
  $f[\#]$ conveys identity, if $f[x]=f[y]$ whenever $x=y$.
    \label{defConveys}
\end{definition}
\begin{eqnarray*}
    x&=&y\\\downmapsto&f&\downmapsto\\f[x]&=&f[y]
\end{eqnarray*} 
After $y=f[x]!$
\begin{eqnarray*}
    y&=&f[x]\\
    \downmapsto&\operatorname{Depth}&\downmapsto\\1&\neq&2
\end{eqnarray*}
$\operatorname{Depth}[\#]$ does not convey identity.

\begin{definition}
  A \emph{function} $f$ from a set $A$ to a set $B$ relates every
  element $x\in A$ to one element $y\in B$, or no one.  Let
  $\mathbf{F}[A,B]$ be the set of functions from $A$ to $B$.
    \label{defFunction}
\end{definition}
Actually, (def.  \ref{defFunction}) is so informal that a formal
system can hardly support it.  What \mma\ readily supports is pure
function, $\operatorname{Function}$,
\begin{equation*} 
  f[\#]\&=\operatorname{Function}[x,\ f[x]],\ f[\#]\&[x]\mapsto f[x],\
  \text{a function}\ f\ \triangleright\ f[\#]\&\ \triangleright\ f.
\end{equation*}

\begin{definition}
  For $f\in\mathbf{F}[A,B]\ \triangleright\ f[\#]\&$, and $x\in A$,
  related to $y\in B$ by $f$, $y=f[x]!$
    \label{defExpFun}
\end{definition}
\begin{theorem}
  $f[\#]\&$ expresses a function iff (if and only if) $f[\#]$ conveys
  identity.
    \label{theConveys}
\end{theorem}
\begin{proof}
  Only if ($\Rightarrow$): a set, as opposed to a pattern, does not
  depend on syntax (formalities); $x,x^{*},x=x^{*}$, expressing the
  same element of $A$, are related to $y=f[x]=f[\new{x}]$.  If
  ($\Leftarrow$): construct the quotient of $f$ modulo identity.
\end{proof}
As $\#[[1]]\&=\operatorname{Part}[\#,1]\&$ \cite{Wolfram} expresses a
projection, $\#[[1]]$ conveys identity. Conversely, as
$\operatorname{Depth}[\#]$ does not convey identity,
$\operatorname{Depth}$ (syntax-sensitive) expresses no function.

\begin{definition}
  $f$ depends on $x$ or $x$ is implicit in $f$, $\partial_{x}f\neq0$,
  if $f$ matches $g\_[x]_{s}$.
    \label{defDependenceImplicit}
\end{definition}
No formal system can completely support $g\_[x]_{s}$.  \mma\ does not
support semantic functional patterns, neither \emph{Semantica},
because $\operatorname{Solve}$ does not support functional equations.
For $f$ depending on $x$, $f=g[x]$, $f[x]$ is not a value of the
function $f$, but a value of a function $h,\ h[x]=g[x][x]$.
\begin{definition}
  $x$ is completely explicit in $f[x]$ if $f$ does not depend on $x$.
    \label{defExplicit}
\end{definition}
    
Formalization (to express the elements without the sets), carried out
to algebra, yields formal algebra.
In particular, some basic (relational) rules of pure function algebra
are
\begin{eqnarray}
    (f\_+g\_)[x\_]&=&f\_[x\_]+g\_[x\_]!
    \\
    (f\_g\_)[x\_]&=&f\_[x\_]g\_[x\_]!
    \\ 
    x\_\circ y\_=\operatorname{Circle}[x\_,y\_]!\
    f\_\circ g\_[x\_]&=&f\_[g\_[x\_]]!
    \label{forFunAlg}
\end{eqnarray}

As for a vector space, in order to reuse existing rules, while
discerning field elements, I propose to express, on one hand, both
internal and external multiplicative laws by $\operatorname{Times}$,
orderless,
\[ 
x\_y\_=\operatorname{Times}[x\_,y\_]!\ x\_y\_=y\_x\_!
\] 
on the other hand, a field element by $x\in K!$
\begin{eqnarray*}
  -1\in K!&&x\_-y\_= x\_+(-1)y\_!  
\end{eqnarray*}
Formal function algebra is specialized in formal function $K$-algebra
by
\[
f\_[\_]/;f\in K\rightarrow f.
\]
\begin{definition}
  A function $f$ is a $K$-linear , $f\in\lin[K]$, if
    \begin{eqnarray*}
        f[x\_+y\_]=f[x\_]+f[y\_],&&
        f[\lambda\_x\_]=\lambda\_f[x\_]/;\lambda\in K.
    \end{eqnarray*}
\end{definition}
For $f\in\lin[K],\ f[x\_-y\_]=f[x\_]-f[y\_]$. 

\subsection{Perturbation}

\begin{definition}
  The perturbation and error symbols are $\operatorname{Star},\delta$.
  For $x,x^{*}$ expressing elements of an additive group,
  $x\triangleright x^{*}$, with the error
  \[
  \delta[x]=x^{*}-x.
  \]
  For $x,x^{*}$ expressing elements of a real vector space, and a
  perturbation amplitude $\epsilon\in[0,1]$, $x\triangleright
  x^{*}_{\epsilon}$, with the error
  \[
  \delta_{\epsilon}[x]=x^{*}_{\epsilon}-x.
  \]
  \begin{xalignat*}{2}
    x^{*}&=\operatorname{Star}[x],& 
    x^{*}_{\epsilon}&=\operatorname{Star}[\epsilon,x],
    \\
    &&\delta_{\epsilon}[x]&=\delta[\epsilon,x],
    \\
    x^{*}_{0}&\rightarrow x,&x^{*}_{1}&\rightarrow x^{*},
    \\
    f\_[x]^{*}&=f\_^{*}[x^{*}],&f\_[x]^{*}_{\epsilon}&=
    f\_^{*}_{\epsilon}[x^{*}_{\epsilon}].
  \end{xalignat*}
  \begin{itemize}
  \item $x$ is \emph{constant} (non-variable), $x\in\cst$, if
    $\delta[x]=\delta_{\_}[x]=0$.
    
  \item $x$ is \emph{unshielded} (non-shielded), $x\in\uns$, if
    $\delta_{\epsilon\_}[x]=\epsilon\_\delta[x]$.
  \end{itemize}
    \label{defPerturbation} 
\end{definition}
$\delta_{\epsilon}[x]=\epsilon\delta[x]$ for $\epsilon\in\{0,1\}$ and
even for $\epsilon\in[0,1]$, if $x\in\uns$.

From (def.  \ref{defPerturbation}), we infer 
\begin{xalignat*}{2}
  \delta_{0}[\_]&\rightarrow0,&\delta_{1}&\rightarrow\delta,
  \\
  x\_^{*}/;x\in\cst&\rightarrow x,&
  x\_^{*}_{\epsilon}/;x\in\cst&\rightarrow x.
\end{xalignat*}
As $\cst,\uns$ express no sets outside the formal system, I call
them pseudo-sets.  
If $x\in\cst$, then $x\in\uns$.  The formal system should at least
partly support the pseudo-inclusion $\cst\subset\uns$, as in
\[
x\in\cst!\ x\in\uns\mapsto\operatorname{True}.
\]

The interest of $x\triangleright x^{*}$ without $x^{*}\rightarrow x$
is when $x^{*}$ matches a simple pattern, that $x$ does not match.
Typically, $x\in\mathbb{R}$ and $x^{*}$ is a floating-point numeral,
matching $\_\operatorname{Real}$.
%
%

$\delta[\#]$ is not meant to convey identity.  For example, let $x,y$
be the lengths of both branches of the upright-faced letter
\textsf{V}; although ideally $x=y$, there are drawing errors
$\delta[x],\delta[y]$, that need not be identical to zero nor to each
other.
\begin{eqnarray*}
    x&=&y\\\downmapsto&\delta&\downmapsto\\\delta[x]&\neq&\delta[y]
\end{eqnarray*}
A physicist would say (privately, so that only another physicist can
understand): perturbation breaks the Curie principle. From (the.
\ref{theConveys}), $\delta$ expresses no function.  If you reject the
pure function $\delta$, then you must consider $\delta x$ as an atom,
and you cannot speak of error, abstractly or symbolically.

Perturbation and error expressions are naturally and informally
abbreviated using total perturbation and total error (like ``total
derivation'') $\Delta,\star$ (tab. \ref{tabChaRul}).

\begin{table}
    \caption{compound error (chain rule)}
    \label{tabChaRul}
    \[\begin{array}{ccccc}
                           & &(f+\delta_{\epsilon}[f])
                           [y+\delta_{\epsilon}[y]]
    \\
                           & &\downeq
    \\
    \delta_{\epsilon}[f[x]]&=&f[x]^{*}_{\epsilon} &-&f[x]
    \\
    \triangledown          & &\triangledown       & &\triangledown
    \\
    \Delta_{\epsilon}[f]   &=&f^{\star}_{\epsilon}&-&f
    \end{array}\]
\end{table}
\begin{theorem}
    \[
    \Delta_{\epsilon}[f]\triangleleft\delta_{\epsilon}[f[x]]=
    f[x^{*}_{\epsilon}]-f[x]+\delta_{\epsilon}[f][x^{*}_{\epsilon}].
    \]
    \label{theChaVar}
\end{theorem}
\begin{proof}
  Use (\ref{forFunAlg}).
\end{proof}

Like motion, perturbation is relative to a steady frame, made of
constant symbols:
\begin{equation} 
    \operatorname{Plus}\in\cst!\
    \operatorname{Times}\in\cst!\
    -1\in\cst!\ 
    \operatorname{Power}\in\cst!\
    \operatorname{Circle}\in\cst!
    \label{cstFrame}
\end{equation}

\begin{theorem}
  For $f\in\cst$, $\operatorname{Star},f$ commute,
    \begin{eqnarray*}
        f[x]^{*}_{\epsilon}=f[x^{*}_{\epsilon}],&&
        \delta_{\epsilon}[f[x]]=f[x^{*}_{\epsilon}]-f[x].
    \end{eqnarray*}
    If moreover $f\in\lin[K]$, then $\delta,f$ commute, 
    \[
    \delta_{\epsilon}[f[x]]=f[\delta_{\epsilon}[x]].
    \]
    \label{theCst<>Star}
\end{theorem}

\begin{theorem}[product error]
    \[
    \delta_{\epsilon}[yx]=y\delta_{\epsilon}[x]+
    x^{*}_{\epsilon}\delta_{\epsilon}[y].
    \]
    If moreover $x,y\in\uns$, then
    \[
    \delta_{\epsilon}[yx]=
    \epsilon(y\delta[x]+x\delta[y])+\epsilon^{2}\delta[x]\delta[y].
    \]
    \label{theVarPro}
\end{theorem}
\begin{proof}
  Use (the.  \ref{theCst<>Star}), with
  $f\rightarrow\operatorname{Times}$; $\operatorname{Times}\in\cst$
  (\ref{cstFrame}) and $\operatorname{Times}$ is orderless.
\end{proof}
\begin{theorem}
  For $x,y\in\cst$, $x+y,xy,x^{-1}\in\cst$, $\cst$ is an algebra.
  Moreover, $x^{y},x[y],x\circ y\in\cst$.
  $\operatorname{Star}\in\lin[\cst],\ \delta\in\lin[\cst]$.  $\cst$ is
  the kernel of $\delta$.  $\uns$ is a $\cst$-vector space.  If
  $f\in\uns$ and $x\in\cst$, then $f[x]\in\uns$.  $\uns$ is stable for
  $f\in\cst,\ f\in\lin[K]$. 
    \label{theUnsSta}
\end{theorem}
\begin{proof}
  Firstly, use (\ref{cstFrame}).  Lastly, use (the.
  \ref{theCst<>Star}),
  \[
  \forall x\in\uns,\ \delta_{\epsilon}[f[x]]=f[\delta_{\epsilon}[x]]
  =f[\epsilon\delta[x]]=\epsilon f[\delta[x]]=\epsilon\delta[f[x]],\ 
  f[x]\in\uns.
  \]
\end{proof}
Attention: from (the.  \ref{theVarPro}), the product of unshielded
expressions is in general shielded, $\uns$ is not an algebra.

\section{Experiment}

\subsection{Observability}

\begin{definition}
  The world is a real vector space $\mathcal{T}$.  A \emph{state} is
  $T\in\mathcal{T}$. For $p\in\mathbb{N}$,
    \[
    \mathcal{T}_{p}\rightarrow\operatorname{If}[p=0,\mathcal{T},
    \mathbf{F}[\mathbb{R}^{p},\mathcal{T}]].
    \]
    A \emph{material system} is $T\in\mathcal{T}_{p}$.  If $p\neq0$,
    then an \emph{input} of $T$ is $z\in\mathbb{R}^{p},\ \exists
    T[z]\in\mathcal{T}$.  The input number of $T$ is $p$.
\end{definition}
\begin{definition}
  For a material system $T$, an expression $x$ is abstract
  (non-concrete), $x\in\abs$, if it does not depend (def.
  \ref{defDependenceImplicit}) on $T$.
    \label{defAbstract}
\end{definition}
The complexity of concrete expressions decreases with concrete
specialization (def.  \ref{defSpecialization}).
\begin{definition}
  For $n,p\in\mathbb{N},\ 
  M\in\mathbf{F}[\mathcal{T}_{p},\mathbb{R}^{n}],\ 
  T\in\mathcal{T}_{p}$, the \emph{material property} $M[T]$ is
  \emph{observable}, as the output of the \emph{experiment} $\{M,T\}$,
  and $n$ is the output number of the \emph{experiment}, if $M[T]$ is
  calculable.  For $p=0$, $M$ is a state function and the experiment
  is static.
  \label{defExperiment}
\end{definition}
The experiment aims at partly evaluating $T$, which is not required to
be observable a priori.
Let $M$ be a state function: for $T\in\mathcal{T}$, $\{M,T\}$ is a
static experiment; for $p\neq0,\ T\in\mathcal{T}_{p}$,
$\{M\circ\#\&,T\}$ is an experiment, and its output is the function
$M\circ T$.

\begin{definition}
  For $n,p\in\mathbb{N},\ 
  M\in\mathbf{F}[\mathcal{T}_{p},\mathbb{R}^{n}],\ 
  T\in\mathcal{T}_{p}$, $\Gamma$ is a \emph{gauge} of $M[T]$, sensitive
  to the \emph{flux} $\Phi[T]$, if $M[T]=\Gamma[\Phi[T]]$ and $\Gamma$
  is calculable.
    \label{defGauge}
\end{definition}
$\Gamma$ aims at making $\Gamma[\Phi[T]]$ less complex that $\Phi[T]$,
which is not required to be observable a priori.  $\Gamma,\Phi$ express laws of
physics (or any experimental science).
\begin{theorem}
  For $M[T]\in\mathbb{R}^{n}$, $M[T]$ observable, and
  $P\in\mathbf{F}[\mathbb{R}^{n},\mathbb{R}]$, $P$ calculable,
  $P[M[T]]$ is observable, as the the output of the experiment
  $\{P\circ M,T\}$.
    \label{theCalObs}
\end{theorem}
\begin{proof} 
  The compound of calculable functions is calculable.
\end{proof}
For example, $n\rightarrow 1,\ f\rightarrow\log$, if $M[T]>0$, then
$\log[M[T]]$ is observable.
\begin{definition}
  For an experiment $\{T,M\}$, and a calculable function $P$ of
  $M[T]$, $P$ interprets $M[T]$ to $\new{M}[T]=P[M[T]]$.  The
  interpretation $P$ is \emph{tautologic} if $P=\id$, \emph{private}
  if $\new{M}\notin\abs$.
    \label{defInterpretation}
\end{definition}
$\#[[1]]\&$ interprets $\{M_{1}[T],M_{2}[T]\}$ to $M_{1}[T]$, yielding
a new experiment $\{M_{1},T\}$.
\begin{theorem}
  For two experiments $\{M,T\}$ and $\{\new{M},T\}$, and a calculable
  function $P$ of $M[T]$, $P$ interprets $M[T]$ to $\new{M}[T]$ iff
  $P$ is a gauge of $\new{M}[T]$, sensitive to $M[T]$.  Moreover, for
  $M\in\abs$, the interpretation $P$ from $M[T]$ is private (def.
  \ref{defInterpretation}) iff the gauge $P$ of $\new{M}[T]$ is
  concrete (def. \ref{defAbstract}).
  \label{theGauge<->interpretation}
\end{theorem}
\begin{proof}
  Use (def.  \ref{defGauge}, def.  \ref{defInterpretation}, the.
  \ref{theCalObs}) and $\new{M}[T]=P[M[T]]$.
\end{proof}
A private interpretation should be presented as a concrete gauge.

\subsection{Controllability}

\begin{definition}
  Let $S\in\mathbf{F}[\mathbf{F}[\mathbb{R},\mathbb{R}],\mathbb{R}]$,
  $S$ calculable.  The set of material systems $T$,
  \emph{controllable} by the servo-function $S$, according to the
  state function $R\in\mathbf{F}[\mathcal{T},\mathbb{R}]$, is
    \begin{multline} 
      \mathcal{C}[R,S]=\{T\in\mathcal{T}_{1},\ 
      \exists!\rho[T]\in\mathbb{R},\ R[T[\rho[T]]]=0,\\
      R\circ T\ \text{observable},\ \rho[T]=S[R\circ T]\}.
    \end{multline}
    For $T\in\mathcal{C}[R,S]$, the \emph{eigeninput} of $T$ is
    $\rho[T]$.
    \label{defEigeninput}
\end{definition}   
For $T\in\mathcal{T}_{1}$, $S$ interprets $R\circ T$ to $\rho[T]$.
For example, let $S[f]$ be the root of
$f\in\mathbf{F}[\mathbb{R},\mathbb{R}]$, $f$ existing everywhere,
continuous, strictly monotonous and crossing $0$.  $S[f]$ is
calculable by dichotomy.  For $T\in\mathcal{T}_{1}$,
$T\in\mathcal{C}[R,S]$ if $R\circ T$ is like $f$ before.

\begin{definition}
    The constraint pure function is $C,\ \ul{x\_}=C[x\_]$, 
    \begin{equation*}
        \ul{(M\_[T\_][\#]\&)_{s}}/;
        (T\in\mathcal{C}[R,S],M[T]\notin\mathcal{C}[R,S])
        \rightarrow M[T][\rho[T]].
    \end{equation*}
    \label{defConstraint}
\end{definition}
\begin{theorem}
  For $P\in\abs$ (def.  \ref{defAbstract}), $C,P$ commute.
    \label{theC<>Abs}
\end{theorem}
\begin{proof}
    \[
    \ul{P[M[T][\#]\&]}=P[M[T][\rho[T]]]=P[\ul{M[T][\#]\&}].
    \]
    For $M[T]$ matching semantically a function, like the derivative
    $M[T]=T'$, shortly,
    \[
    \ul{P[M[T]]}=P[M[T][\rho[T]]]=P[\ul{M[T]}].
    \]
\end{proof}

\subsection{Perturbation under constraint}

\begin{theorem}
  If $\rho\in\cst$ and $\operatorname{Star}\in\abs,\ \epsilon\in\abs$,
  then $C,\operatorname{Star}$ commute on $M[T]$: perturbation under
  constraint is the same as constraint under perturbation.
    \label{theC<>Star}
\end{theorem}
\begin{proof}
  Thanks to $\rho\in\cst$ (in the lower horizontal branch) and
  $\operatorname{Star}\in\abs,\ \epsilon\in\abs$ (in the right
  vertical branch),
    \begin{eqnarray*}
        M[T]&
        \overset{\operatorname{Star}[\epsilon,\#]\&}{\mapsto}&
        M^{*}_{\epsilon}[T^{*}_{\epsilon}]
        \\
        C\downmapsto&&C\downmapsto
        \\
        M[T][\rho[T]]&
        \overset{\operatorname{Star}[\epsilon,\#]\&}{\mapsto}&
        M^{*}_{\epsilon}[T^{*}_{\epsilon}][\rho[T^{*}_{\epsilon}]]
    \end{eqnarray*} 
\end{proof}
Moreover, $C\in\cst$ iff $\rho\in\cst$ if $S\in\cst,\ R\in\cst$ (def.
\ref{defEigeninput}). To satisfy the hypothesis of (the.
\ref{theC<>Star}), 
\begin{gather}
  \operatorname{Star}\in\abs!\ \delta\in\abs,\ \epsilon\in\abs!
    \label{StarAbs}\\
    S\in\cst!\ R\in\cst!\ \rho,C\in\cst.
    \label{cstFrame1}
\end{gather}

Theorem \ref{theC<>Star} is implicitly used in thermodynamics, as a
principle, asserting that, in a state to state transformation, like a
monothermal compression, the final value of a state function, like
energy, does not depend on the order of perturbation (compression) and
constraint (constant temperature).  $C\in\cst$ means that compression
does not affect the reservoir temperature.

If $\operatorname{Plus}\in\abs,\ R\in\abs$, then, considering that
$\operatorname{Plus},\ R\circ T$ semantically match functions,
\begin{eqnarray*}
    \ul{M_{1}[T]+M_{2}[T]}=\ul{M_{1}[T]}+\ul{M_{2}[T]},&&
    \ul{R\circ T}=R[\ul{T}]
\end{eqnarray*}
and, with perturbation,
\begin{equation*}
    (\ul{R\circ T})^{*}_{\epsilon}=\ul{(R\circ T)^{*}_{\epsilon}}=
    \ul{R\circ T^{*}_{\epsilon}}=
    R\circ T^{*}_{\epsilon}[\rho[T^{*}_{\epsilon}]]=
    R[\ul{T^{*}_{\epsilon}}]\triangleright
    \ul{R}^{\star}_{\epsilon}.
\end{equation*}

\begin{table}
    \caption{constraint and perturbation shorthands}
    \label{tabCStar}
\[\begin{array}{l|l|l}
    x\_            &\ul{x}=              &\ul{x}\triangleright\\\hline
                   &\rho[T]              &\rho\\
    T              &T[\rho]              &\ul{T}\\
    T'             &T'[\rho]             &\ul{T}'\\
                   &\rho[T^{*}_{\epsilon}]
                                         &\rho^{\star}_{\epsilon}\\
    T^{*}_{\epsilon}          
                   &T^{*}_{\epsilon}[\rho^{\star}_{\epsilon}]
                                         &\ul{T}^{*}_{\epsilon}\\
    R\circ T^{*}_{\epsilon}
                   &R[\ul{T^{*}_{\epsilon}}]     
                                         &\ul{R}^{\star}_{\epsilon}\\
    \delta[T]      &\delta[T][\rho]      &\ul{\delta[T]}
\end{array}\]
\end{table}

Let $T,T^{*}_{\epsilon}\in\mathcal{C}[R,S],\ 
\delta[T]\notin\mathcal{C}[R,S]$.  By (def.  \ref{defConstraint}),
\begin{eqnarray*}
    T^{*}&=&T+\delta[T]\\
    \downmapsto&C&\downmapsto\\
    T^{*}[\rho[T^{*}]]&\neq&
    (T+\delta[T])[\rho[T]]=T^{*}[\rho[T]]
\end{eqnarray*}
$\ul{\#}$ does not convey identity, $C$ expresses no function (the.
\ref{theConveys}).  Equivalently, $\ul{T^{*}-T}$ matches the semantic
pattern of (def.  \ref{defConstraint}) in two ways, $T\_\rightarrow T$
or $T\_\rightarrow T^{*}$ (a donkey problem again).

\section{Design}

\subsection{Complexity reduction}

Let $T\in\mathcal{T}$.  The user-experimentalist wants to observe (to
make observable) not the whole of $T$, but some less complex material
property $\sigma[T]\in\mathbb{R}$.  He finds out an input in $T$, on
which $\sigma[T]$ depends,
\[
\exists\operatorname{Function}
[\epsilon\in[0,1],T^{*}_{\epsilon}\in\mathcal{T}]\triangleright
T^{*}_{\#}\&\in\mathcal{T}_{1},\ T=T^{*}_{0},\ 
\partial_{\epsilon}\sigma[T^{*}_{\epsilon}]\neq0,
\]
and a state function $R\in\mathbf{F}[\mathcal{T},\mathbb{R}]$, such
that $R[T^{*}_{\epsilon}]$ is observable for $\epsilon\in[0,1]$, hence
the synthetic experiment $\{(R\circ\#)\&,T^{*}_{\#}\&\}$.  The problem
is, to find an interpretation $P$ from $R\circ T^{*}$ to $\sigma[T]$,
yielding the experiment $\{\sigma[\#[0]]\&,T^{*}_{\#}\&\}$, of output
$\sigma[T^{*}_{0}]=\sigma[T]=P[R\circ T^{*}]$.  The pure function
$\#[0]\&$ expresses the Dirac measure, thus linked to analytic
experiment.

I assume a linear gauge $\sigma[T]\#\&$ of $R[T]$, sensitive to
$\Phi[T]$, and
\[
\sigma\in\lin[\mathbb{R}],\ \sigma\in\cst,\ 
\delta[\sigma[T]]=-\sigma[T],\ T\in\uns.
\]
\begin{eqnarray*}
    \forall\epsilon\in[0,1],\
    \sigma[T^{*}_{\epsilon}]&=&\sigma[T]+\epsilon\sigma[\delta[T]]=
    \sigma[T](1-\epsilon),
    \nonumber\\
    R[T^{*}_{\epsilon}]&=&\sigma[T]\Phi[T^{*}_{\epsilon}](1-\epsilon),
    \label{Runs}\\
    (R\circ T^{*})'[0]&=&\partial_{\epsilon}R[T^{*}_{\epsilon}]
    /.\epsilon\rightarrow0
    \nonumber\\
    &=&\sigma[T](d\Phi[T][\delta[T]]-\Phi[T]).
    \nonumber
\end{eqnarray*}
If $d\Phi[T][\delta[T]]=0$ and $\Phi[T]$ is observable, then
$-\Phi[T]^{-1}\#\&$ interprets $(R\circ T^{*})'[0]$ to $\sigma[T]$, as
expected.

In the case \cite{AlbaredeWeighing}, $\sigma[T]$ cannot be so easily
evaluated; $\sigma[T]$ is a component of $T$:
\begin{gather*}
  T=\{\sigma[T],\new{T}\},\ \new{T}\in\cst!
  \\
  \delta[T]=\{-\sigma[T],0\},\ 
  T^{*}_{\epsilon}=\{\sigma[T](1-\epsilon),\new{T}\}.
  \\
  \forall\epsilon\in[0,1],\ R[T^{*}_{\epsilon}]=
  \sum_{\tim=0}^{\infty}r_{\tim}[\sigma[T],\new{T}]\epsilon^{\tim}.
\end{gather*}
Instead of $\sigma[T]$, $r_{\tim}[\sigma[T],\new{T}],\ \tim=0,1\ldots$
are observable (after another interpretation, that is interpolation).
The user has not only an inverse problem, but also the problem that
$r_{\tim}[\#,\new{T}]\&$ is a private interpretation from $\sigma[T]$.

The signal to noise ratio is
$r_{1}[\sigma[T],\new{T}]/r_{0}[\sigma[T],\new{T}]$; the shielding
ratio is $r_{2}[\sigma[T],\new{T}]/r_{1}[\sigma[T],\new{T}]$ (null if
$\Phi\in\uns$, from (\ref{Runs})).  The user wants to minimize the
latter and to maximize the former, while avoiding large variations of
$R$ (for private reasons).  Therefore, he relaxes $\new{T}\in\cst$ and
finds an input in $\new{T}$, on which $R[T]$ depends,
\begin{gather*}
  \exists\new{T}^{*}_{\#}\&,\ \new{T}=\new{T}^{*}_{0},\ 
  \partial_{\epsilon}R[\{\sigma[T],\new{T}^{*}_{\epsilon}\}]\neq0,
  \\
  T_{2}[x,y]=\{\sigma[T](1-x),\new{T}^{*}[y]\},\ 
  \partial_{x,y}T_{2}[x,y]=0,
  \\
  \new{R}[x\_]=R[x\_]-R[T],
\end{gather*}
such that $R[T_{2}[x,y]]$ is observable for $x,y\in[0,1]$, hence the
experiment $\{\new{R}\circ\#\&,T_{2}\}$.  For
$T_{2}[x,\#]\&\in\mathcal{C}[\new{R},S]$ (def. \ref{defConstraint}),
$\new{R}\circ T_{2}$ is interpreted by
$S[\operatorname{Function}[y,\#[x,y]]]\&$ to $\rho[T_{2}[x,\#]\&]$:
\begin{equation*}
    \rho[T_{2}[x,\#]\&]=S[\new{R}\circ T_{2}[x,\#]\&]
    =S[\operatorname{Function}[y,\new{R}\circ T_{2}[x,y]]].
\end{equation*}
Synthetically, $\new{R}\circ T_{2}$ is interpreted by
$\operatorname{Function}[x,S[\operatorname{Function}[y,\#[x,y]]]\&]$
to $\operatorname{Function}[x,\rho[T_{2}[x,\#]\&]]$.
\begin{eqnarray*}
    \forall\epsilon\in[0,1],\ \rho[T_{2}[\epsilon,\#]\&]=
    \sum_{\tim=0}^{\infty}\new{r}_{\tim}[\sigma[T],\new{T}^{*}_{\#}\&]
    \epsilon^{\tim}.
\end{eqnarray*}
Instead of $\sigma[T]$,
$\new{r}_{\tim}[\sigma[T],\new{T}^{*}_{\#}\&],\ \tim=0,1\ldots$ are
observable, without large variations of $R$, but with a worse privacy
problem, since the private variable is now functional.

Complexity constraints leads to observe another material property than
initially contemplated; ignoring this fact is a frequent cause of
misunderstanding.

Experiment design follows from calculability.  According to (def.
\ref{defExperiment}) and \cite[3.1.2]{BarrereMma}, $\sigma[T]$ is
observable if the formal system can evaluate
$\operatorname{step}_{0}[T]$,
\begin{xalignat*}{2}
  \operatorname{step}_{\tim\_}[T]/;
  \operatorname{test}_{\tim}[\new{\sigma}_{\tim}[T]]&\rightarrow\tim,&
  \operatorname{step}_{\tim\_}[T]&\rightarrow
  \operatorname{step}_{\tim+1}[T],
  \\
  \new{\sigma}_{\tim\_}[T]&\rightarrow
  P_{\tim}[\sigma,T][\new{\sigma}_{\tim-1}[T]],&
  \operatorname{test}_{\tim\_}[T]&\rightarrow
  P_{\tim}[\operatorname{test},T][\operatorname{test}_{\tim-1}[T]].
\end{xalignat*}
This experimental program is such that $\tim$ is incremented only after
the test has failed, so that $\operatorname{step}_{0}[T]$ returns the
minimum $\tim$ passing the test (also a fixed point of the pure function
$\operatorname{step}$).  For $\tim\in\mathbb{N}\pined$,
$P_{\tim}[\sigma,T]$ is a gauge of $\new{\sigma}_{\tim}[T]$, sensitive
to $\new{\sigma}_{\tim-1}[T]$, or an interpretation from
$\new{\sigma}_{\tim-1}[T]$ to $\new{\sigma}_{\tim}[T]$, public if
$\partial_{T}P_{\tim}[\sigma,T]=0$.

The user is responsible for the initial idea $\new{\sigma}_{0}[T]$,
for ruling $P$ and for interrupting.  He plays a game against nature
(as in \cite{Albarede421}) and $P$ expresses his strategy.  For
example, if $\sigma[T]$ is the average of a random variable, then
$P_{\tim}[\sigma,\#]$ means simply ``try again'',
\[
P_{\tim}[\sigma,T][\new{\sigma}_{\tim-1}[T]]\rightarrow
\operatorname{Random}[T],
\]
while $P_{\tim}[\operatorname{test},\#]$ should be some convergence
criterion.

$\partial_{\tim}P_{\tim}\neq0$ allows dynamic programming, or
debugging: to have $\operatorname{step}_{0}[T]$ evaluated step by step
and to decide $P_{\tim}$ as late as possible, taking into account past
experiments, that is \emph{experience}.

\subsection{Error reduction}

If $\sigma[T]$ was already known (numerically), then the user could
test accuracy by
\begin{equation*}
    \operatorname{test}_{\tim}[x\_]
    \rightarrow(\absolute{x-\sigma[T]}\le e\_\operatorname{Real}).
\end{equation*}
Actually, $\sigma[T]$ is not usually known.  $\operatorname{test}$ may
ensure termination, but not accuracy: this all the ambiguity of
``finishing a job''.  When experimenting has stopped at $\tim$, the
error is, symbolically,
\[
\new{\sigma}_{\tim}[T]-\sigma[T]=\sigma^{*}[T^{*}]-\sigma[T]=
\sigma[T^{*}]-\sigma[T]+\delta[\sigma][T^{*}].
\]
The realization error is $\sigma[T^{*}]-\sigma[T]$; the programming
error is $\delta[\sigma][T]\approx\delta[\sigma][T^{*}]$.  Good
realization is ruined by bad programming, and conversely; both errors
ought to be simultaneously small.  

\section{Conclusion}

\subsection{Rhetoric of experiment}

The understanding of dependence, or semantic pattern matching, is
essential in experiment, while no formal system can completely support
it. Now that we have got a rather precise language for dependence
(tab.  \ref{tabDependence}) and experiment (tab.
\ref{tabExperiment}), let us play with it, for public, non-technical,
expert-free discussion.

\begin{table}
\caption{the lexicon of dependence and independence}
\label{tabDependence}
\begin{center}\begin{tabular}{llll}
    non-universal& universal    \\
    specialized  & unspecialized\\
    implicit     & explicit     \\     
    concrete     & abstract     \\ 
    private      & public       \\
    variable     & constant     \\
    shielded     & unshielded
\end{tabular}\end{center}
\end{table}
\begin{table}
\caption{the lexicon of experiment}
\label{tabExperiment}
\begin{center}\begin{tabular}{llll}    
    constraint    & controllable  & eigeninput     & error         \\     
    experiment    & exp. program  & gauge          & flux          \\     
    input         & interpretation                                 \\    
    mat. property & mat. system   & observable     & output        \\
    perturbation  & servo-function& state          & state function
\end{tabular}\end{center}
\end{table}

Dependence is not only relation but action, or calculability.
Relation without action easily leads to non-sense.
\begin{itemize}
\item An electrical engineer needs to evaluate a current $i$, from the
  heat $ri^{2}$ released by a resistor.  The initial experiment is
  $\{r\#^{2},i\}$.  Assuming that $r$ is observable (which is another
  problem), $ri^{2}$ is interpreted by $(r^{-1}\#)^{1/2}\&$ to
  $\absolute{i}$.  The sign of the current cannot be determined
  (unless some other kind of gauge is used).  But $ri^{2}$ also
  matches $(k\_i)_{s}$, with $k\rightarrow ri$, and $ri^{2}$ is
  interpreted by $k^{-1}\#\&$ to $i$.  This interpretation fails,
  because $k$ is not observable.
    
\item More generally, any observable $M[T]$ cannot be interpreted to
  $\sigma[T]$ by $P,\ P[M[T]]=\sigma[T]$, unless $P$ is calculable.
\end{itemize}

Implicit inputs can play tricks on the user-experimentalist.
\begin{itemize}
\item A heating engineer needs to evaluate a temperature $\theta$ in a
  furnace.  He uses as a gauge a thermocouple, the output of which is
  interpreted by an amplifier $P$.  Unfortunately, the amplifier is
  directly heated by the furnace, so that its gain depends on
  $\theta$.  $P$ is a private interpretation from the thermocouple
  output or the gauge of some material property, hardly related to
  temperature.  This can be corrected, by either insulating the
  amplifier or expliciting temperature in gain.
  
\item Two surveyors, one working near the North pole, the other near
  the equator, need to evaluate the sum of angles of triangular
  oceanic plots.  They may disagree, until they understand that the
  laws of geometry, expressed by $\Phi$, depend on curvature,
  decreasing with latitude.  This is a typical problem of hidden
  variable, or restrictive semantic pattern matching.
    
\item Hidden variables have been similarly suspected in quantum
  mechanics, in particular by Einstein.
\end{itemize}

To name a gauge after the material property it was intended to
evaluate, instead of the material property it actually evaluates, is
to take our desires for realities. Without interpretation, we can hope
to observe nothing but natural numbers, as in sheep counting, and,
hopefully, real numbers, as read on rules.
\begin{itemize}
\item An accelerometer does not evaluate acceleration, but
  displacement, to be interpreted to acceleration by second
  derivation.
  
\item A flow meter does not evaluate flow rate, but some concrete
  material property, like a turbine angular velocity, hence
  ``calibration'' problems.
    
\item A hygrometer does not evaluate humidity, but the length of a
  hair.  Of course, hair-length meters would not sell as well as
  hygrometers.
  
\item A lie detector\ldots
\end{itemize}

\subsection{Universality, not reductionism}

Many formal systems seem to occur naturally (tab.  \ref{tabSpeForSys})
and to behave similarly.  However, their universal reduction (def.
\ref{defUniForSys}) is far from being effective, because of thick
complexity barriers.
Universality does not mean reductionism.
\begin{table}
\caption{formal systems?}
\label{tabSpeForSys}
\begin{center}\begin{tabular}{l|lll}
              &$T$                    &$\Phi[T]$      &$R[\Phi[T]]$\\\hline
    universal &rules, data            &trace          &output\\    
    quantum   &potential, mass        &wave function  &probability\\
    transport &source, cross section  &particle flux  &average number\\
    mechanical&initial condition, mass&motion         &position\\
    human     &intention, knowledge   &thought        &speech\\
    economic  &needs, goods           &market         &price\\
\end{tabular}\end{center}
\end{table}

%
%
%
%
%
%

\section*{Thanks}

Thanks to R\'emi Barr\`ere, for many fruitful discussions, about
mathematics, computer science and philosophy.

\bibliographystyle{unsrt} \bibliography{biblio}
\end{document}